\documentclass[11pt, a4paper, onecolumn, metastyle]{snowflake-paper}

\makeatletter
\renewcommand\bibentry[1]{\nocite{#1}{\frenchspacing\@nameuse{BR@r@#1\@extra@b@citeb}}}
\makeatother

\usepackage{kantlipsum, lipsum}
\usepackage{snowflake-colors}
\usepackage{float}
\usepackage{booktabs} 
\usepackage{multicol} 
\usepackage{graphicx} 
\usepackage{adjustbox} 
\usepackage{cleveref}
\usepackage{subcaption}
\usepackage{multirow} 
\usepackage{array}
\usepackage{listings}
\usepackage{wasysym}
\usepackage{tabularx,ragged2e,adjustbox}
\newcolumntype{Y}{>{\RaggedRight\arraybackslash}X}

\usepackage{xspace}
\usepackage{amssymb}

\newcommand{\EmptyCircle}{%
  \tikz[baseline=-0.6ex]\draw[line width=0.4pt] (0,0) circle (0.9ex);
}
\newcommand{\FullCircle}{%
  \tikz[baseline=-0.6ex]\fill (0,0) circle (0.9ex);
}
\newcommand{\HalfCircle}{%
  \tikz[baseline=-0.6ex]{
    \fill (0,0) -- (90:0.9ex) arc[start angle=90,end angle=270,radius=0.9ex] -- cycle;
    \draw[line width=0.4pt] (0,0) circle (0.9ex);
  }
}
\newcommand{\midtool}{\texttt{MidTool}}
\newcommand{\midtooldata}{\texttt{MidTool-Mix}}
\definecolor{lightgray}{gray}{0.95}
\definecolor{midtool}{RGB}{235,245,255}
\definecolor{lightgreen}{RGB}{235,250,235}

\usepackage{pgfplots}
\pgfplotsset{width=10cm, height=6cm, compat=1.18}

\usepackage[authoryear, sort&compress, round]{natbib}

\usepackage{subcaption}
\usepackage{cleveref}
\usepackage[most, breakable]{tcolorbox}
\usepackage{amsmath}
\usepackage{fontawesome5} 
\usepackage{pifont}

\usepackage{makecell}

\usepackage{algorithm}
\usepackage{algpseudocode}   
\usepackage{tikz}
\usepackage{enumitem}

\usepackage{pgf}             
\usepackage{multirow}
\usepackage[version=4]{mhchem}

\newcommand{\cmark}{\ding{51}}  
\newcommand{\xmark}{\ding{55}}  

\usepackage{twemojis} 
\tcbuselibrary{skins, breakable, listings, theorems}
\usepackage{xparse} 

\NewDocumentCommand{\scorecell}{m g}{%
  \pgfmathparse{int(#1*100*0.25)}%
  \xdef\pct{\pgfmathresult}%
  \cellcolor{red!\pct!white}{%
    #1%
    \IfNoValueTF{#2}{}{ $\pm$ #2}%
  }%
}

\newcolumntype{P}[1]{>{\raggedright\arraybackslash}p{#1}}

\newtcolorbox{promptbox}[1][]{
    promptstyle,
    title=Prompt,
    #1
}
\tcbset{
    promptstyle/.style={
        enhanced,
        colback=white,
        colframe=black,
        colbacktitle=gray!20,
        coltitle=black,
        rounded corners,
        sharp corners=north,
        boxrule=0.5pt,
        drop shadow=black!50!white,
        attach boxed title to top left={
            xshift=-2mm,
            yshift=-2mm
        },
        boxed title style={
            rounded corners,
            size=small,
            colback=gray!20
        },
        before upper={\setlength{\parindent}{0pt}},
    },
    systempromptstyle/.style={
        enhanced,
        colback=green!4,
        colframe=black,
        colbacktitle=gray!20,
        coltitle=black,
        rounded corners,
        sharp corners=north,
        boxrule=0.5pt,
        before upper={\setlength{\parindent}{0pt}},
        drop shadow=black!50!white,
        attach boxed title to top left={
            xshift=-2mm,
            yshift=-2mm
        },
        boxed title style={
            rounded corners,
            size=small,
            colback=green!10
        },
        
    },
    replystyleg/.style={
        enhanced,
        colback=green!15,
        colframe=black,
        colbacktitle=green!30,
        coltitle=black,
        boxrule=0.5pt,
        drop shadow=black!50!white,
        rounded corners,
        sharp corners=north,
        attach boxed title to top right={
            xshift=-2mm,
            yshift=-2mm
        },
        boxed title style={
            rounded corners,
            size=small,
            colback=green!40
        }
    },
    replystyler/.style={
        enhanced,
        colback=red!15,
        colframe=black,
        colbacktitle=red!40,
        coltitle=black,
        boxrule=0.5pt,
        drop shadow=black!50!white,
        rounded corners,
        sharp corners=north,
        attach boxed title to top right={
            xshift=-2mm,
            yshift=-2mm
        },
        boxed title style={
            rounded corners,
            size=small,
            colback=red!40
        }
    },
    thought/.style={
        enhanced,
        colback=orange!15,
        colframe=black,
        colbacktitle=orange!40,
        coltitle=black,
        boxrule=0.5pt,
        drop shadow=black!50!white,
        rounded corners,
        sharp corners=north,
        attach boxed title to top right={
            xshift=-2mm,
            yshift=-2mm
        },
        boxed title style={
            rounded corners,
            size=small,
            colback=orange!40
        }
    }
}

\usepackage{caption}  
\tcbuselibrary{listings}      
\tcbuselibrary{breakable}     
\newtcblisting{PromptFloat}{
  enhanced, listing only, breakable,
  float*=htb,
  colback=white, colframe=black,
  title style={colback=gray!20},
  sharp corners=north, drop shadow=black!50!white,
  boxrule=0.5pt, fonttitle=\sffamily\bfseries,
  left=2mm,right=2mm,top=1mm,bottom=1mm,
  listing options={basicstyle=\small\ttfamily,
                   breaklines=true,
                   breakautoindent=false,
                   breakindent=0pt,
                   columns=fullflexible,
                   frame=none,
                   xleftmargin=0pt,        
                   inputencoding=utf8},    
}

\definecolor{abs}{HTML}{eef8fd} 
\definecolor{urllink}{HTML}{11567F} 

\graphicspath{{figures/}}
\title{\midtool: Mid-training Data Synthesis for Agentic Tool Use}

\paperurl{}
\reportnumber{} 

\theoremstyle{definition}

\author[1,$\dagger$]{Fengqing Jiang}
\author[2]{Yite Wang}
\author[2]{Boyi Liu}
\author[3,$\dagger$]{Zhaoyang Wang}
\author[2]{Canwen Xu}
\author[2]{\\Zhewei Yao}
\author[1,$\ddagger$]{Radha Poovendran}
\author[2,$\ddagger$]{Yuxiong He}

\affil[1]{University of Washington}
\affil[2]{Snowflake}
\affil[3]{University of North Carolina at Chapel Hill\qquad\qquad\qquad\qquad\qquad}

\affil[$\dagger$]{Work done at Snowflake}
\affil[$\ddagger$]{Co-advising}

\begin{abstract}
Mid-training is increasingly recognized as a critical stage for shaping the capabilities of large language models. Recent work has shown that targeted mid-training can strengthen reasoning-intensive abilities such as math and science, and can also improve agentic capabilities in software-engineering settings. In this work, we study the parallel but less explored agentic capability: general tool use. We present \midtool, an open corpus construction pipeline for agentic tool-use mid-training that combines large-scale web, PDF, and code data with synthesized supervision from real-world tool APIs, MCP skills, and document-grounded workflows. \midtool~is designed to teach models how to recognize tool affordances, ground arguments from context, compose tool call workflow, and recover from incomplete information. We mid-train \texttt{Qwen3-4B-Base} and \texttt{Qwen3-8B-Base} on \midtooldata, and then apply follow-up post-training with both supervised fine-tuning and reinforcement learning. Compared with baselines, \midtooldata~consistently improves downstream performance under both SFT and RL on BFCL, $\tau^2$-Bench, and MCP Universe. These results suggest that general tool use, like other important LLM capabilities, benefits from dedicated mid-training rather than being left entirely to post-training.

\vspace{1em}
\textbf{Contact:} \url{yite.wang@snowflake.com}

\textbf{Data \& Model:} \url{https://hf.co/collections/MidTool/midtool-release}
\end{abstract}
\begin{document}

\maketitle

\begin{figure}[!ht]
    \centering
    \includegraphics[width=0.97\linewidth]{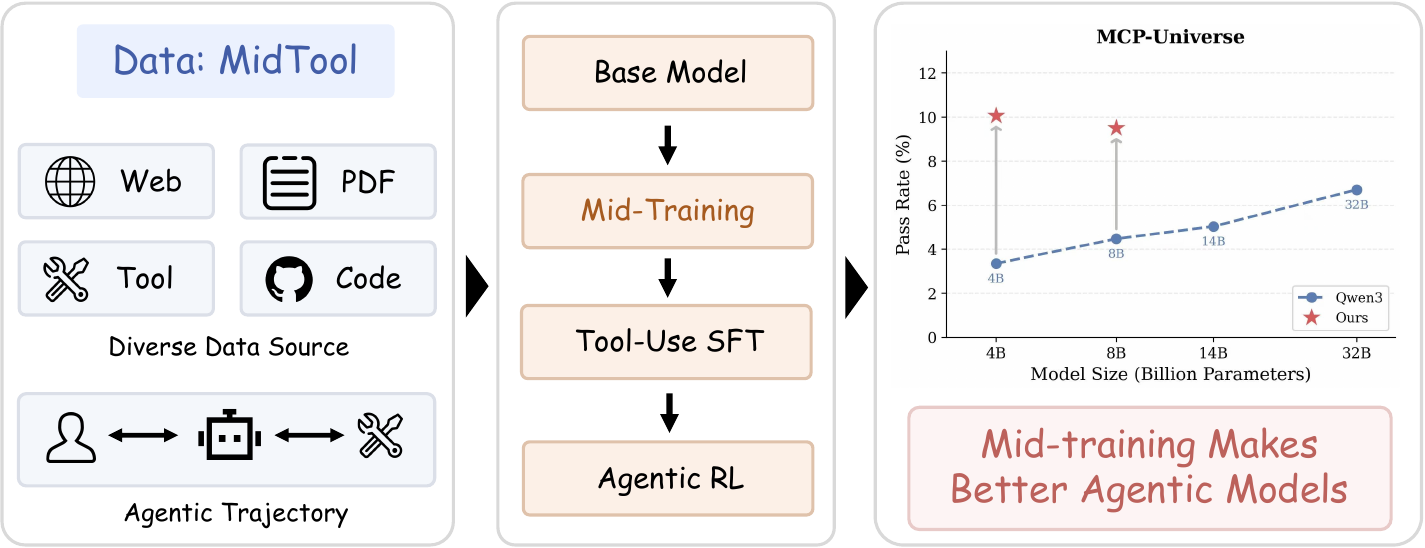}
    \caption{\textbf{Left:} \midtooldata~is a 20.3B-token mid-training corpus built from web, PDF, code, and tool sources with synthesized agentic trajectories. \textbf{Middle:} Training pipeline from base model. \textbf{Right:} On MCP-Universe, our mid-trained 4B and 8B models outperform Qwen3 official models, showing that dedicated mid-training yields stronger agentic capability than scaling alone.}
    \label{fig:teaser}
\end{figure}
\section{Introduction}
Tool use is becoming a defining capability of large language model (LLM) agents.
Strong systems must decide when external tools are needed, ground calls in tool schema, extract arguments from long and noisy context, compose multiple tools into workflows, and recover when information is incomplete.
These behaviors now matter across function calling, API orchestration, interactive agent benchmarks, and emerging Model Context Protocol (MCP) ecosystems \citep{schick2023toolformer, liu2024apigen, bfcl, luo2025mcp}.

Most recent progress on tool use, however, has come from post-training.
Supervised fine-tuning (SFT) and reinforcement learning (RL) on curated traces have substantially improved function calling and agent behavior \citep{song2024agentbank, liu2025toolacewinningpointsllm, prabhakar2025apigen, xu2025toucansynthesizing15mtoolagentic}, but they also place a heavy burden on post-training.
The model must simultaneously acquire a set of atomic agentic capabilities, including tool recognition, schema-grounded argument construction, clarification under missing information, and multi-step execution, from comparatively narrow supervision.
More fundamentally, the knowledge that underlies successful tool use does not live only in explicit trajectories.
It is distributed across developer documentation, manuals, PDFs, code repositories, API specifications, and structured tool definitions, most of which never appear as clean agent demonstrations \citep{qin2023toolllm, liu2024apigen, xu2025toucansynthesizing15mtoolagentic}.

This raises a natural question: \textbf{can general tool-use ability be shaped earlier through dedicated mid-training, rather than being left almost entirely to post-training?} Mid-training is a distinct stage of multi-stage pre-training, bridging the gap between general pre-training and post-training
stages \citep{midtrainsurvey}.
Recent work has shown that targeted mid-training can materially improve reasoning-intensive capabilities \citep{wang2025octothinker, longcat-flash-thinking}, and emerging efforts have begun to explore agentic mid-training for deep research \citep{su2025scalingagentscontinualpretraining, team2025tongyi}, coding \citep{zeng2025glm45, zai2026glm5, xiao2026mimo}, and software engineering (SWE) \citep{kimidev,  davincidev}.
Yet general tool use remains underexplored as a mid-training target.
Compared with math or SWE tasks, tool use data requires covering a broader and more heterogeneous capability surface, including natural-language documentation, executable code patterns, structured schemas, multi-tool workflows, and failure cases caused by missing information.

In this paper, we introduce \midtool, a scalable pipeline for constructing agentic mid-training data of general tool use, together with \midtooldata, the resulting 20.3B-token mixture.
Our pipeline begins from four complementary source families: web pages, PDFs, code repositories, and structured tool artifacts such as APIs and MCP skills.
It then converts these sources into training supervision through two synthesis branches that target the two core deficits of tool use: grounding and execution.
The first branch, \textbf{context-grounded trajectory augmentation}, addresses grounding by turning documentation and code into supervision for recognizing tool boundaries, inferring parameters, and recovering workflow structure from messy real-world artifacts.
The second branch, \textbf{native agentic trajectory synthesis}, addresses execution by constructing executable trajectories directly from real APIs, MCP skills, and collected rollout traces, teaching multi-turn planning, clarification, and recovery with explicit validation of schema grounding, turn order, required arguments, and tool-response consistency.
The resulting mixture is designed to teach models not only what tools exist, but also the atomic agentic capabilities needed to recognize their affordances, compose them into workflows, and recover when information is incomplete. Table~\ref{tab:corpus-comparison} situates our work among representative open corpora and mid-training efforts.
Relative to prior open efforts, our goal is neither a general-domain mixture nor a domain-specific corpus for math, deep research, or software engineering, but a dedicated agentic corpus for \emph{general tool use}.

\begin{table}[!bt]
    \centering
    \resizebox{\textwidth}{!}{%
\begin{tabular}{lcrccc c c}
\toprule
Work & Domain & \makecell{Data \\ Size} & Train & Data Sources & \makecell{Tool \\ Div.} & \makecell{\small Agent \\ \small Traj.} & \makecell{\small Public \\ \small Access}\\
\midrule
FineWeb \citep{fineweb} & General & 15T & PT & Web & \EmptyCircle & \xmark & \cmark\\
Dolmino \citep{olmo2025olmo3} & General & 100B & MT  & Crawl sourced + Synthesized & \EmptyCircle & \xmark & \cmark\\
MegaMath-Web-Pro \citep{wang2025octothinker} & Math & 100B & MT  & Math, QA, Instruction + Synthesized & \EmptyCircle & \xmark & \cmark\\
AgentFounder \citep{su2025scalingagentscontinualpretraining} & \small Deep Research &  300B & MT & Web + Synthesized & \HalfCircle & \cmark & \xmark \\
daVinci-Dev \citep{davincidev} & SWE & 73.1B & MT & Github PR + Synthesized & \HalfCircle & \cmark & \cmark\\ \midrule
\midtooldata~(ours) & Tool Use & 20.3B & MT  & Web, PDF, Code, Tool + Synthesized  & \FullCircle & \cmark & \cmark \\
\bottomrule
\end{tabular}%
}
\begin{flushleft}
\footnotesize
\quad \textit{Note:} \EmptyCircle\ low/none, \HalfCircle\ partial, and \FullCircle\ high; \cmark\ and \xmark\ denote yes and no, respectively.
\end{flushleft}
\vspace{-1em}
    \caption{Comparison of representative work on pre-training (PT) and mid-training (MT) corpora. Tool Div. = tool diversity and Agent Traj. = (include) agentic trajectories. }
    \label{tab:corpus-comparison}
\end{table}

We evaluate whether such mid-training provides value beyond standard post-training by mid-training \texttt{Qwen3-4B-Base} and \texttt{Qwen3-8B-Base} on \midtooldata, followed by the same downstream SFT and optional RL recipes.
Across BFCL \citep{bfcl}, $\tau^2$-Bench \citep{barres2025tau}, and MCP-Universe \citep{luo2025mcp}, we observe a consistent pattern: \midtooldata~improves downstream tool-use performance over SFT-only baselines, and RL usually compounds these gains.
The gains are especially pronounced on harder multi-turn and interactive settings.

\textbf{Contributions.}
Our main contributions are threefold: \textbf{(1)} To our knowledge, we introduce \midtool, the first open pipeline and mid-training dataset designed for general tool use, and construct \midtooldata, a 20.3B-token mixture that combines web, PDF, code, and tool artifacts with both context-grounded augmentation and native agentic trajectories.
\textbf{(2)} We show that dedicated tool-use mid-training consistently improves downstream general tool-use performance for 4B and 8B models across three benchmarks under both SFT and RL, suggesting that mid-training provides a stronger and more stable substrate for subsequent post-training.
\textbf{(3)} We show that \midtooldata~helps reveal a meaningful capability boundary: its benefits are strongest for broad tool-use capability and transfer, while more specialized exploratory behaviors remain distinct. This suggests that capabilities such as deep search likely require dedicated mid-training data beyond a purely general tool-use mixture, providing concrete guidance for future agentic mid-training study.

\section{\midtool: Scalable Pipeline for Agentic Mid-training Data Synthesizing}
 
In this section, we present our pipeline to construct the large-scale mid-training corpus, including data source collection, data preprocessing and agentic trajectory synthesis stages.

\subsection{Stage 1: Data Source Collection}
\begin{figure}[!t]
\vspace{-1em}
    \centering
    \includegraphics[width=\linewidth]{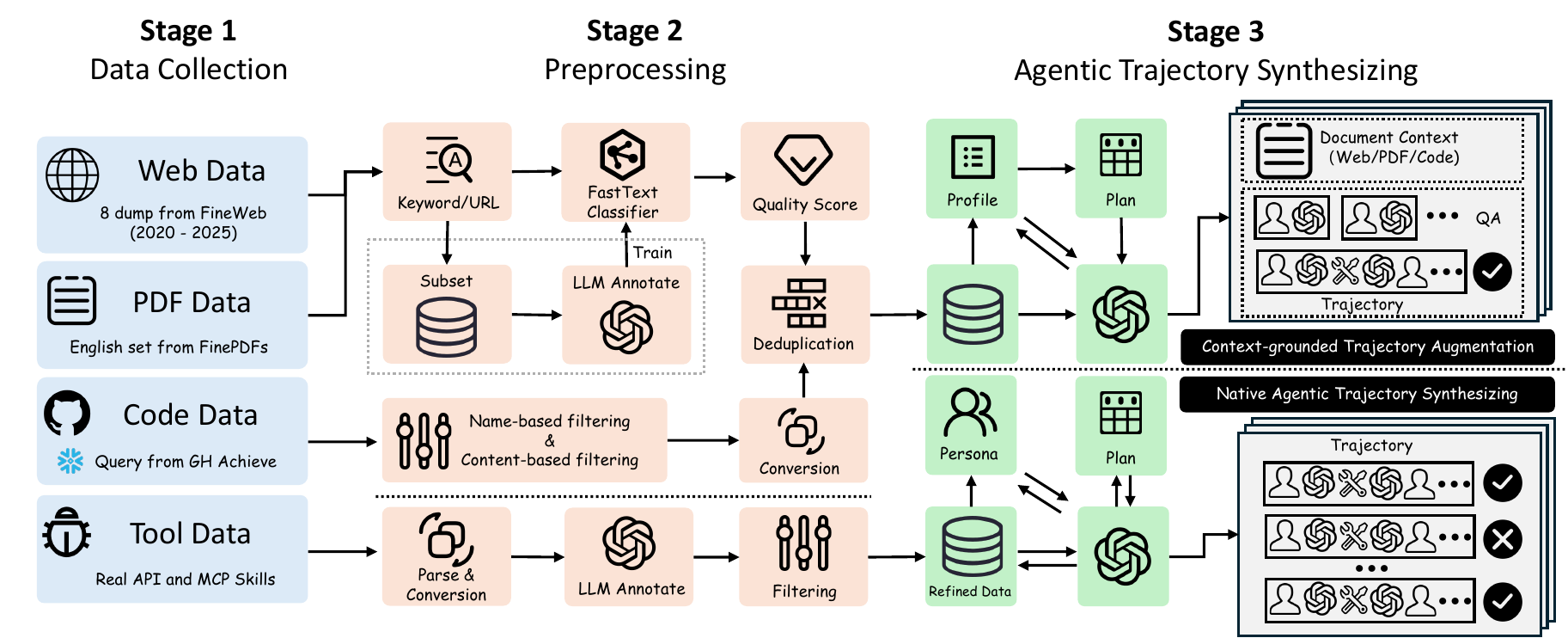}
    \caption{Overview of \midtool~pipeline. Stage 1 collects four complementary source families, including web pages, PDFs, code repositories, and structured tool artifacts. Stage 2 applies source-specific preprocessing, quality control, and deduplication. Stage 3 converts the refined corpus into executable supervision through two branches: context-grounded trajectory augmentation from web/PDF/code documents, and native agentic trajectory synthesis from real APIs and MCP skills.}
    \label{fig:pipeline}
    \vspace{-1em}
\end{figure}

As illustrated in Figure \ref{fig:pipeline}, our data collection stage is designed to cover the complementary signals required for agentic tool use. Effective tool-use behavior depends not only on explicit trajectories, but also on broad technical knowledge, tool-facing documentation, executable code patterns, and grounded tool schemas. We therefore collect four source families: web documents, PDFs, code repositories, and structured tool artifacts.

\textbf{Web data.}
We use processed Common Crawl dumps from FineWeb \citep{fineweb} as the primary large-scale web source. We sample multiple dumps spanning 2020 to 2025 to capture recent technical content, including API references, developer documentation, troubleshooting pages, tutorials, and CLI-style instructions. Relative to later trajectory synthesis, the web corpus provides broad coverage of tool-related concepts, terminology, and workflow descriptions at scale.

\textbf{PDF data.}
Many high-value tool-use resources, such as manuals, product handbooks, and platform documentation, are distributed as PDFs rather than clean HTML pages. We therefore use FinePDFs \citep{kydlicek2025finepdfs} as a complementary crawl source and keep only the English subset. Compared with web pages, PDFs contribute longer-form procedural content and documentation that is often absent from standard web snapshots, but they also require stricter downstream filtering because extraction noise is substantially higher.

\textbf{Code data.}
Tool use is tightly coupled with software artifacts, so we collect repository data from two complementary GitHub slices. We initialize repository discovery from Snowflake GH Archive event data \citep{gharchive}, then apply task-specific filtering. The first slice targets agent-related and MCP-related repositories obtained from GitHub event data, with filtering for recency, activity, and selective licensing. This slice captures concrete tool interfaces, orchestration patterns, and emerging agent engineering practice. The second slice targets high-quality public repositories with strong community signals and recent activity in major programming ecosystems such as Python, Java, JavaScript, TypeScript, Go, Rust, C/C++, C\#, SQL, Shell, and Dockerfile. We further retain repositories that are likely to contain useful libraries, SDKs, frameworks, examples, or educational technical content, while removing personal projects, forks, and benchmark or dataset repositories that would add noise or leakage risk. In particular, we explicitly exclude known benchmark and evaluation repositories from the GitHub code slice through an actively maintained blacklist during data collection; more details are provided in Appendix~\ref{appx:decontamination}.

\textbf{Tool data.}
Finally, we separately collect structured tool artifacts, including REST APIs and MCP skills, to support native agentic trajectory synthesis. Unlike web, PDF, and code sources, these artifacts expose executable schemas, parameter structures, and tool boundaries directly. They therefore provide the most faithful substrate for constructing grounded tool calls, multi-tool plans, and recovery behaviors in the later synthesis stage.

Taken together, these four source families intentionally trade off breadth and structure. Web and PDF data provide broad technical context, code repositories contribute executable patterns and developer workflows, and tool artifacts supply explicit schemas and callable interfaces. The subsequent preprocessing and synthesis stages convert this heterogeneous raw corpus into a unified training mixture for agentic mid-training.

\subsection{Stage 2: Data Preprocessing}

\textbf{Code.} Following \cite{li2023starcoder} and \cite{weber2024redpajama}, we apply multi-phase filtering to construct a high-value subset for effective training. We adapt the extension-based filtering list to exclude files such as binary data, model weights, and logs. We then adapt the heuristics from StarCoder \citep{li2023starcoder} for content-based filtering, including line count, average line length, maximum line length, and alpha ratio. We also convert Jupyter Notebook files to Python text files. For deduplication, we remove exact duplicates by SHA-256 hashing over normalized text, and we remove near-duplicates with MinHash LSH. For accepted high-quality repositories, we apply a filter that matches only documentation-like directories such as docs, examples, tutorials, guides, samples, and cookbook.

\textbf{Web and PDF data.} 
For web and PDF midtraining sources, we use a four-phase pipeline: high-recall keyword/URL prescreening, a lightweight fastText \citep{fasttext} classifier trained on LLM-labeled seed data, document-level quality filtering, and MinHash LSH deduplication. The pipeline is designed to retain developer-oriented documentation and technical reference material while suppressing generic web noise and PDF extraction artifacts, with stricter thresholds for PDFs. Full details are provided in Appendix~\ref{appx:web-pdf-pipeline}.

\subsection{Stage 3: Agentic Trajectory Synthesizing}
As shown in our later study, effective tool-use mid-training must address two complementary deficits.
The first is \emph{grounding}: models often fail to infer tool boundaries, required arguments, and workflow structure from messy real-world artifacts such as documentation, PDFs, and code.
The second is \emph{execution}: even when tool schemas are available, models still struggle to plan across multiple turns, request missing information, sequence calls correctly, and recover from incomplete interactions.
We therefore instantiate a two-branch synthesis pipeline that converts both unstructured documents and structured tool artifacts into normalized agent trajectories.

\textbf{Context-grounded Trajectory Augmentation.}
Many web, PDF, and code documents expose tool affordances and workflows without containing explicit interaction traces, so we use them as the grounding-oriented branch.
  We first apply a lightweight keyword-based prefilter, then use \texttt{Qwen3-235B-A22B-Instruct-2507} to annotate each remaining document with quality score, and a structured affordance
  profile.
  The affordance profile records whether tool responses can be inferred, together with evidence about schema/API structure, code or CLI usage, workflow structure, tool topology, and domain terminology.
  Documents below a quality threshold are kept without augmentation, while for the retained documents, a rule-based planner converts the extracted affordances into a synthesizing plan.

  The planner ties supervision volume to document quality, allocating a bounded budget for diverse QA types and allowing at most one multi-turn chain per document.
  The regular QA decomposes tool use into atomic agentic capabilities, such as tool selection, schema-grounded parameter extraction, format-constrained calls, workflow recognition, and multiple/parallel use, while the trajectory samples cover sequential execution, parameter clarification, tool switching, and long-context reasoning. We use \texttt{Qwen3-235B-A22B-Instruct-2507} to synthesize the augmented data. Only QA pairs and trajectories that pass parsing and semantic quality control are merged into the final training mix.

\textbf{Native Agentic Trajectory Synthesis.}
For structured tool sources, including both API groups and MCP skills, we use a second branch to teach execution and planning directly from executable interfaces.
  We first build a tool inventory by grouping related endpoints or skills, parsing tool definitions, and removing low-signal sources with lightweight prescreening.
  We then use \texttt{GPT-5} to assign each source a quality score together with a feasibility profile over trajectory families, including simple single-call use, complex multiple/parallel tool use, and information-missing settings.
  For sources that pass the quality filter, we recover developer-oriented documentation context, normalize the available tools into canonical executable schemas, and apply targeted schema refinement only when argument descriptions are underspecified.
  We next synthesize diverse user personas and let the model propose candidate trajectory plans, but the final allocation is enforced by a deterministic quality-adaptive budget controller that conditions on source quality, tool count, argument structure, and feasibility constraints, with preference for multi-turn trajectories.
  This prevents overproducing trivial samples and shifts generation toward richer multi-turn behaviors for high-capability sources.
  Finally, category-specific generators instantiate the planned trajectories, using a mixture of \texttt{GPT-5}, \texttt{GPT-5.1}, and \texttt{GPT-5.2} for generation.
  The resulting trajectories are strictly validated for turn ordering, schema grounding, required arguments, and tool-response consistency; invalid generations are additionally retried with quality-control feedback before discard.
  We also mix in rollout trajectories collected from the agentic world model (AWM) \citep{wang2026agentworldmodel} synthesized environments, so that the model learns not only isolated atomic capabilities, but also their composition into robust agent behavior. Finally, we also incorporate filtered agentic traces from the Nemotron Agentic dataset \citep{blakeman2025nemotron} to further scale up the native trajectory portion. 

\begin{figure}[t]
\centering
\begin{minipage}[t]{0.56\textwidth}
\centering
\captionof{table}{Data mixture statistics. Token counts are in billions. Slash-separated values denote source corpus / context-grounded augmentation.}
\label{tab:data-mixture}
\small
\setlength{\tabcolsep}{4pt}
\begin{tabular}{l r r r}
\toprule
Source & Tokens (B) & Samples & Ratio \\
\midrule
Web               & 4.4 / 4.1 & 6.86M & 42\% \\
PDF               & 2.6 / 2.1 & 1.34M & 23\% \\
Code              & 3.8 / 1.5 & 2.60M & 26\% \\
\midrule
\small Native Agentic Trajectory & 1.8 & 0.42M & 9\% \\
\midrule
Total             & 20.3 & 11.22M & 100\% \\
\bottomrule
\end{tabular}
\end{minipage}\hfill
\begin{minipage}[t]{0.38\textwidth}
\vspace{-5pt}
\centering
\includegraphics[width=\linewidth,page=1]{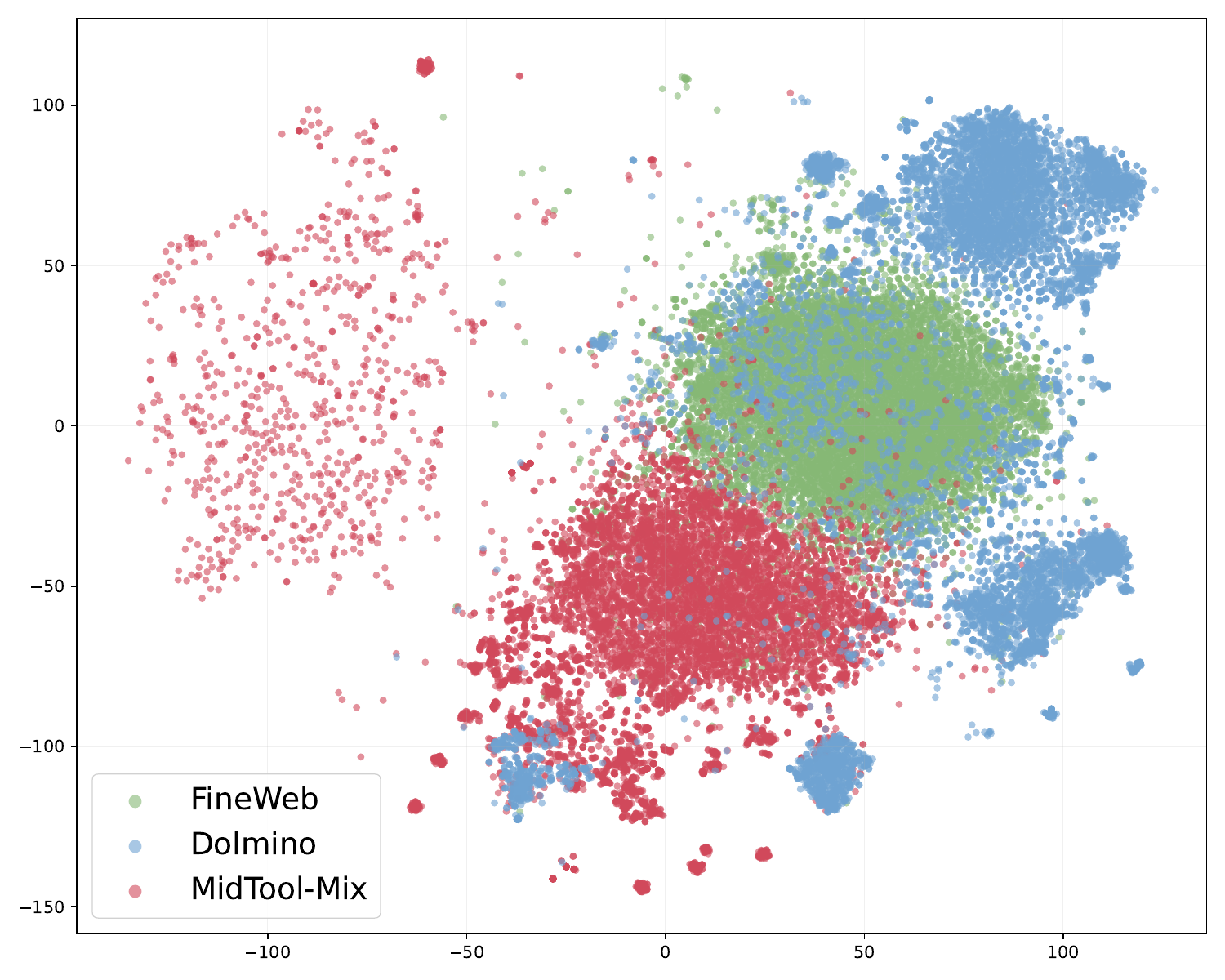}
\captionof{figure}{t-SNE visualization.}
\label{fig:tsne-dataset}
\end{minipage}
\end{figure}

\subsection{Analysis of \midtooldata}
\textbf{\midtooldata.} Our final training mixture consists of three components: (1) a high-quality source mixture spanning filtered web, code, and PDF corpora; (2) context-grounded trajectories, including QAs, derived from those sources; and (3) native agentic trajectories, including synthesized trajectories built from real corpora, collected interaction rollouts, and filtered Nemotron Agentic dataset. Following \citet{olmo2025olmo3}, we normalize all trajectories into a plain chat-style template without special control tokens such as \texttt{<im\_start>}.

Table \ref{tab:data-mixture} summarizes the resulting composition. In total, \midtooldata~contains 20.3B tokens and 11.22M samples. For web, PDF, and code, we report the retained source tokens and the associated context-grounded augmentation separately using slash-separated counts. The overall mixture is intentionally balanced across web (42\%), code (26\%), and PDF (23\%) sources, while native agentic trajectories contribute an additional 9\% of the total budget. This allocation preserves broad technical coverage for grounding from natural documents, while reserving a non-trivial fraction of the mixture for executable agent supervision. Appendix~\ref{appx:composition} provides a finer-grained breakdown by augmentation type, trajectory shape, and tool inventory.

Figure \ref{fig:tsne-dataset} provides a qualitative comparison against baseline pre-training dataset FineWeb and mid-training mix Dolma 3 Dolmino (Dolmino) by \citet{olmo2025olmo3}. The visualization comes from 2K sampled examples per dataset using the embedding model \texttt{Arctic-Embed-2.0-L} \citep{yu2024arcticembed20multilingualretrieval}. \midtooldata~marginally overlaps with both reference corpora, which is expected because our work still draws on broad technical web data. Regarding the distinct region of Dolmino, we believe it is associated with the effort on math/science reasoning-focused corpus in Dolmino, which is not our primary focus.  At the same time, \midtooldata~occupies distinct regions that are not covered well by either dataset, indicating that our pipeline does not merely recover a narrow subset of generic pretraining and dedicated mid-training data. Instead, it shifts the distribution toward documentation-heavy, workflow-oriented, and agentic tool-use content. We additionally audit \midtooldata~for overlap with our evaluation benchmarks and find no evidence of leakage (Appendix~\ref{appx:decon}).

\section{Experiment}
\subsection{Setup}

\textbf{Baselines and Training Setup.}
 Unlike prior work that studies tool use primarily as a post-training objective, our pipeline starts from pretrained base models and injects tool-use capability earlier through mid-training. We study two base models, \texttt{Qwen3-4B-Base} and \texttt{Qwen3-8B-Base}, and compare four training recipes for each scale: raw base model with SFT and optional RL; base model mid-trained with \midtooldata, followed by SFT and optional RL. For SFT, we fine-tune the model on a sampled 100K tool-use subset from TOUCAN \citep{xu2025toucansynthesizing15mtoolagentic}. Both mid-training and SFT are conducted with ArcticTraining \citep{snowflake_arctictraining_2026} on 32 H200 GPUs. We adopt the AWM setup~\citep{wang2026agentworldmodel} using 526 synthetic tool-use environments for agentic RL training with 8 B200 GPUs. Detailed hyperparameters are deferred to Appendix \ref{appx:hyperparameters}. We also compare to the post-trained \texttt{Qwen3-4B} and \texttt{Qwen3-8B} models, and disable thinking to align with our setting.

\textbf{Benchmarks.}
We evaluate general agentic tool-use capability on three complementary benchmarks: v3 split of BFCL (BFCLv3) \citep{bfcl}, verified $\tau^2$-Bench \citep{barres2025tau, cuadron2025saber}, and MCP-Universe \citep{luo2025mcp}. BFCL measures function-calling quality across both single-turn and multi-turn settings, and is particularly useful for isolating grounded tool selection, argument construction, and hallucination behavior. $\tau^2$-Bench stresses interactive task completion in realistic verticals such as airline, retail, and telecom, making it a useful probe of multi-step execution and recovery. MCP-Universe evaluates execution over real MCP servers spanning domains such as browser automation, finance, location, and web search, and therefore serves as a harder test of out-of-distribution tool generalization. 
We follow the eval harness developed by \cite{wang2026agentworldmodel}.

\subsection{Main Results}
Tables \ref{tab:bfcl-results}, \ref{tab:tau2-results}, and \ref{tab:mcp-universe-results} summarize the main results. Across both model sizes, we observe a consistent pattern: \midtooldata~improves downstream agentic tool-use performance over SFT-only baselines, and RL usually compounds these gains further. The largest improvements typically appear on benchmarks or subsets that require longer interaction horizons, stronger schema grounding, or more robust adaptation to unfamiliar tools. Additional training-dynamics evidence is provided in Appendix~\ref{appx:exp-result}, where we show that \midtooldata~also yields better SFT convergence and faster early-stage RL adaptation. Appendix~\ref{appx:visual-tool-use} also reports an exploratory pilot on visual tool use.

\begin{table*}[t]
\centering
\caption{BFCLv3 results for Qwen3-4B and Qwen3-8B base models. For each scale, we compare SFT and SFT+RL with and without prior mid-training on \midtooldata; released Qwen3 models are included as references. We report single-turn, multi-turn (MF = Missing Function, MP = Missing Parameters, LC = Long Context), hallucination (Hallu.), and overall performance. All values are percentages and higher is better.}
\label{tab:bfcl-results}
\small
\resizebox{\textwidth}{!}{%
\begin{tabular}{lcccccccccc}
\toprule
\multirow{2}{*}{\textbf{Model Setting}} & \multicolumn{2}{c}{\textbf{Single Turn}} & \multicolumn{5}{c}{\textbf{Multi-Turn}} & \multirow{2}{*}{\textbf{Hallu.}} & \multirow{2}{*}{\textbf{Overall}} \\ \cmidrule(lr){2-3} \cmidrule(lr){4-8}
 & \textbf{Non-live} & \textbf{Live} & \textbf{Base} & \textbf{MF} & \textbf{MP} & \textbf{LC} & \textbf{Avg.} &  &  \\
\midrule

Qwen3-4B & 39.58\% & 24.35\% & 10.50\% & 9.00\% & 6.00\% & 10.00\% & 8.88\% & 9.05\% & 24.27\% \\ \midrule
\rowcolor{lightgray}
Qwen3-4B-Base + SFT & 59.94\% & 43.75\% & 21.50\% & 14.00\% & 10.00\% & 16.50\% & 15.50\% & \textbf{60.46\%} & 39.73\% \\ 
\rowcolor{lightgray}
Qwen3-4B-Base + SFT + RL & 59.65\% & 39.90\% & 27.00\% & 18.00\% & 13.00\% & 18.00\% & 19.00\% & 55.19\% & 39.51\% \\ 
\rowcolor{lightgreen}
Qwen3-4B-Base + \midtooldata~+ SFT & 66.38\% & 57.74\% & \textbf{36.00\%} & 23.00\% & 19.50\% & 28.00\% & 26.63\% & 56.95\% & 50.25\% \\
\rowcolor{lightgreen}
Qwen3-4B-Base + \midtooldata~+ SFT + RL & \textbf{76.44\%} & \textbf{58.48\%} & 34.50\% & \textbf{24.50\%} & \textbf{20.50\%} & \textbf{31.00\%} & \textbf{27.63\%} & 60.10\% & \textbf{54.18\%} \\
\midrule \midrule
Qwen-8B & 30.73\% & 26.13\% & 31.00\% & 22.50\% & 15.00\% & 21.50\% & 22.50\% & 9.77\% & 26.45\% \\ \midrule
\rowcolor{lightgray}
Qwen3-8B-Base + SFT & 66.40\% & 51.22\% & 30.50\% & 26.50\% & 16.00\% & 28.00\% & 25.25\% & \textbf{65.03\%} & 47.62\% \\
\rowcolor{lightgray}
Qwen3-8B-Base + SFT + RL & 68.52\% & 39.60\% & 43.00\% & 22.00\% & 19.00\% & 33.00\% & 29.25\% & 52.79\% & 45.79\% \\
\rowcolor{lightgreen}
Qwen3-8B-Base + \midtooldata~+ SFT & 65.73\% & \textbf{55.37\%} & 38.00\% & 31.50\% & 24.00\% & 35.50\% & 32.25\% & 59.82\% & 51.12\% \\
\rowcolor{lightgreen}
Qwen3-8B-Base + \midtooldata~+ SFT + RL & \textbf{72.58\%} & 55.14\% & \textbf{50.50\%} & \textbf{33.50\%} & \textbf{25.00\%} & \textbf{41.50\%} & \textbf{37.63\%} & 64.22\% & \textbf{55.12\%} \\
\bottomrule
\end{tabular}%
}
\end{table*}

\textbf{Mid-training yields capabilities beyond post-training alone.}
As shown in Table \ref{tab:bfcl-results}, adding \midtooldata~before SFT substantially improves overall BFCL at both 4B and 8B scale, with the strongest variants appearing after RL. The gains are especially pronounced on the multi-turn subsets: at 4B, the average multi-turn score rises by more than 10 points over the SFT-only baseline, and the same pattern holds at 8B, where \midtooldata~combined with RL yields the best multi-turn performance. The fact that the largest gains concentrate on harder multi-step settings suggests that mid-training contributes capabilities that are not reliably induced by standard post-training alone, especially stronger grounding of atomic agentic capabilities together with better planning and execution over longer interaction horizons.

\begin{table*}[t]
\centering
\caption{$\tau^2$-Bench results for Qwen3-4B and Qwen3-8B base models. For each scale, we compare SFT and SFT+RL with and without prior mid-training on \midtooldata; released Qwen3 models are included as references. We report Pass@1 and Pass@4 for airline, retail, telecom, and overall; higher score is stronger interactive capacity.}

\label{tab:tau2-results}
\small
\setlength{\tabcolsep}{4pt}
\resizebox{\textwidth}{!}{%
\begin{tabular}{lcccccccc}
\toprule
\multirow{2}{*}{\textbf{Model Setting}} & \multicolumn{2}{c}{\textbf{Airline}} & \multicolumn{2}{c}{\textbf{Retail}} & \multicolumn{2}{c}{\textbf{Telecom}} & \multicolumn{2}{c}{\textbf{Overall}} \\ \cmidrule(lr){2-3} \cmidrule(lr){4-5} \cmidrule(lr){6-7} \cmidrule(lr){8-9}
 & \textbf{Pass@1} & \textbf{Pass@4} & \textbf{Pass@1} & \textbf{Pass@4} & \textbf{Pass@1} & \textbf{Pass@4} & \textbf{Pass@1} & \textbf{Pass@4} \\
\midrule
Qwen3-4B & 21.00\% & 32.00\% & 16.23\% & 35.09\% & 3.51\% & 5.26\% & 11.87\% & 22.30\% \\ \midrule
\rowcolor{lightgray}
Qwen3-4B-Base + SFT & 17.50\% & 34.00\% & 9.43\% & 24.56\% & 3.73\% & 10.53\% & 8.54\% & 20.50\% \\
\rowcolor{lightgray}
Qwen3-4B-Base + SFT + RL & \textbf{31.00\%} & \textbf{52.00\%} & 18.20\% & 39.47\% & 0.00\% & 0.00\% & 13.04\% & 25.54\% \\
\rowcolor{lightgreen}
Qwen3-4B-Base + \midtooldata~+ SFT & 17.00\% & 38.00\% & 20.83\% & 47.37\% & 1.54\% & 4.39\% & 12.23\% & 28.06\% \\
\rowcolor{lightgreen}
Qwen3-4B-Base + \midtooldata~+ SFT + RL & 20.00\% & 40.00\% & \textbf{33.55\%} & \textbf{57.89\%} & \textbf{6.36\%} & \textbf{18.42\%} & \textbf{19.96\%} & \textbf{38.49\%} \\
\midrule \midrule
Qwen-8B & 19.00\% & 44.00\% & 12.06\% & 37.72\% & 5.04\% & 16.67\% & 10.43\% & 30.22\% \\ \midrule
\rowcolor{lightgray}
Qwen3-8B-Base + SFT & 12.00\% & 38.00\% & 15.79\% & 38.60\% & 4.39\% & 13.16\% & 10.43\% & 28.06\% \\
\rowcolor{lightgray}
Qwen3-8B-Base + SFT + RL & 15.50\% & 38.00\% & 30.92\% & 57.02\% & \textbf{5.26\%} & \textbf{19.30\%} & 17.63\% & 38.13\% \\
\rowcolor{lightgreen}
Qwen3-8B-Base + \midtooldata~+ SFT & 18.50\% & 46.00\% & 26.32\% & 61.40\% & 1.54\% & 3.51\% & 14.75\% & 34.89\% \\
\rowcolor{lightgreen}
Qwen3-8B-Base + \midtooldata~+ SFT + RL & \textbf{24.50\%} & \textbf{52.00\%} & \textbf{39.04\%} & \textbf{67.54\%} & 2.19\% & 6.14\% & \textbf{21.31\%} & \textbf{39.57\%} \\
\bottomrule
\end{tabular}%
}
\end{table*}

\textbf{The learned gains transfer to realistic agentic tasks.}
We see this most clearly on $\tau^2$-Bench and MCP-Universe in Table \ref{tab:tau2-results} and \ref{tab:mcp-universe-results}, which are both more realistic and complex than BFCL. On $\tau^2$-Bench, \midtooldata~substantially improves overall performance at both scales, nearly doubling overall Pass@1 at 4B and yielding clear gains in overall Pass@1 / Pass@4 at 8B. The gains are strongest on airline and retail, while telecom remains difficult, but the overall pattern still shows transfer to interactive task completion. The same trend appears on MCP-Universe, where \midtooldata~raises both overall score and pass rate at 4B and 8B, with especially visible gains after RL. Taken together, these two benchmarks suggest that \midtooldata~does not merely improve tool use, but learns a more general prior for interacting with realistic environments and previously unseen tool ecosystems.

\begin{table*}[t]
\centering
\caption{MCP-Universe results for Qwen3-4B and Qwen3-8B base models. For each scale, we compare SFT and SFT+RL with and without prior mid-training on \midtooldata; released Qwen3 models are included as references. We report score and pass rate; higher is better.}
\label{tab:mcp-universe-results}
\small
\setlength{\tabcolsep}{3.2pt}
\resizebox{\textwidth}{!}{%
\begin{tabular}{lcccccccccccc}
\toprule
\multirow{2}{*}{\textbf{Model Setting}} & \multicolumn{2}{c}{\textbf{Browser Auto.}} & \multicolumn{2}{c}{\textbf{Financial}} & \multicolumn{2}{c}{\textbf{Location}} & \multicolumn{2}{c}{\textbf{Multi-server}} & \multicolumn{2}{c}{\textbf{Web Search}} & \multicolumn{2}{c}{\textbf{Overall}} \\ \cmidrule(lr){2-3} \cmidrule(lr){4-5} \cmidrule(lr){6-7} \cmidrule(lr){8-9} \cmidrule(lr){10-11} \cmidrule(lr){12-13}
 & \textbf{Score} & \textbf{Pass} & \textbf{Score} & \textbf{Pass} & \textbf{Score} & \textbf{Pass} &\textbf{Score} & \textbf{Pass} & \textbf{Score} & \textbf{Pass} &\textbf{Score} & \textbf{Pass} \\
\midrule
Qwen3-4B & 19.42 & 5.88\% & 9.17 & 7.50\% & 43.51 & \textbf{2.86\%} & 16.17 & 0.00\% & 0.00 & 0.00\% & 16.05 & 3.35\% \\ \midrule
\rowcolor{lightgray}
Qwen3-4B-Base + SFT & 15.44 & 2.94\% & 5.00 & 5.00\% & 38.98 & 0.00\% & 13.67 & 0.00\% & 0.00 & 0.00\% & 13.20 & 1.68\% \\
\rowcolor{lightgray}
Qwen3-4B-Base + SFT + RL & 17.65 & 5.88\% & 6.67 & 5.00\% & 39.69 & 0.00\% & 17.00 & 0.00\% & 0.00 & 0.00\% & 14.50 & 2.23\% \\
\rowcolor{lightgreen}
Qwen3-4B-Base + \midtooldata~+ SFT & \textbf{21.81} & \textbf{8.82\%} & 14.27 & 12.50\% & 46.81 & \textbf{2.86\%} & \textbf{19.50} & 0.00\% & 0.00 & 0.00\% & 18.66 & 5.03\% \\
\rowcolor{lightgreen}
Qwen3-4B-Base + \midtooldata~+ SFT + RL & 19.73 & \textbf{8.82\%} & \textbf{38.33} & \textbf{37.50\%} & \textbf{49.50} & 0.00\% & 16.17 & 0.00\% & 0.00 & 0.00\% & \textbf{23.80} & \textbf{10.06\%} \\
\midrule \midrule
Qwen-8B & 22.92 & 8.82\% & 12.50 & 12.50\% & 25.78 & 0.00\% & 7.83 & 0.00\% & 0.00 & 0.00\% & 13.06 & 4.47\% \\ \midrule
\rowcolor{lightgray}
Qwen3-8B-Base + SFT & 17.40 & 5.88\% & 7.50 & 7.50\% & 40.52 & 0.00\% & \textbf{20.33} & \textbf{5.00\%} & 0.00 & 0.00\% & 15.18 & 3.35\% \\
\rowcolor{lightgray}
Qwen3-8B-Base + SFT + RL & 13.60 & 2.94\% & 17.50 & 17.50\% & 36.26 & \textbf{2.86\%} & 18.67 & 0.00\% & 0.00 & 0.00\% & 15.67 & 5.03\% \\
\rowcolor{lightgreen}
Qwen3-8B-Base + \midtooldata~+ SFT & 19.12 & 5.88\% & 11.46 & 10.00\% & 48.31 & \textbf{2.86\%} & 19.50 & 0.00\% & 0.00 & 0.00\% & 17.82 & 3.91\% \\
\rowcolor{lightgreen}
Qwen3-8B-Base + \midtooldata~+ SFT + RL & \textbf{24.63} & \textbf{11.76\%} & \textbf{30.83} & \textbf{30.00\%} & \textbf{58.35} & \textbf{2.86\%} & 19.50 & 0.00\% & 0.00 & 0.00\% & \textbf{25.16} & \textbf{9.50\%} \\
\bottomrule
\end{tabular}%
}
\end{table*}

\begin{table*}[t]
\centering
\caption{Ablation across BFCLv3, $\tau^2$-Bench, and MCP-Universe under the Qwen3-4B-Base + SFT setting, where the downstream post-training recipe is fixed and only the mid-training corpus is changed. Rows under \midtool~are additive over the filtered raw sources: \emph{Processed data w/o traj.} mid-trains on the processed raw data only, with no synthesized trajectories; the next two rows each add a \emph{single} synthesis branch on top of it, and are therefore alternatives rather than successive steps; \midtooldata~is the complete mixture combining all. \texttt{Dolmino-20BT} is a matched-budget generic mid-training baseline.
}

\label{tab:ablation-bfcl}
\small
\setlength{\tabcolsep}{4pt}
\resizebox{\textwidth}{!}{%
\begin{tabular}{lcccccccc}
\toprule
\multirow{2}{*}{\makecell{\textbf{Mid-training}\\ \textbf{Data}}} & \multicolumn{4}{c}{\textbf{BFCLv3}} & \multicolumn{2}{c}{\textbf{$\tau^2$-Bench}} & \multicolumn{2}{c}{\textbf{MCP-Universe}} \\ \cmidrule(lr){2-5} \cmidrule(lr){6-7} \cmidrule(lr){8-9}
 & \textbf{Non-live} & \textbf{Live} & \textbf{Multi-turn} & \textbf{Overall} & \textbf{Pass@1} & \textbf{Pass@4} & \textbf{Score} & \textbf{Pass} \\
\midrule
No Mid-training & 59.94\%$\textcolor{gray}{_{+0.0}}$ & 43.75\%$\textcolor{gray}{_{+0.0}}$ & 15.50\%$\textcolor{gray}{_{+0.0}}$ & 39.73\%$\textcolor{gray}{_{+0.0}}$ & 8.54\%$\textcolor{gray}{_{+0.0}}$ & 20.50\%$\textcolor{gray}{_{+0.0}}$ & 13.20$\textcolor{gray}{_{+0.0}}$ & 1.68\%$\textcolor{gray}{_{+0.0}}$ \\
\midrule
\rowcolor{lightgray}
Dolmino-20BT & 61.44\%$\textcolor{ForestGreen}{_{\scriptstyle +1.5}}$ & 51.74\%$\textcolor{ForestGreen}{_{\scriptstyle +8.0}}$ & 16.13\%$\textcolor{ForestGreen}{_{\scriptstyle +0.6}}$ & 43.10\%$\textcolor{ForestGreen}{_{\scriptstyle +3.4}}$ & 7.37\%$\textcolor{BrickRed}{_{\scriptstyle -1.2}}$ & 21.22\%$\textcolor{ForestGreen}{_{\scriptstyle +0.7}}$ & 5.41$\textcolor{BrickRed}{_{\scriptstyle -7.8}}$ & 0.00\%$\textcolor{BrickRed}{_{\scriptstyle -1.68}}$ \\
\midrule
\rowcolor{lightgreen} %
\bf \midtool & & & & & & & & \\
\rowcolor{lightgreen}
\quad Processed data w/o traj. & 60.40\%$\textcolor{ForestGreen}{_{\scriptstyle +0.5}}$ & 52.60\%$\textcolor{ForestGreen}{_{\scriptstyle +8.9}}$ & 13.90\%$\textcolor{BrickRed}{_{\scriptstyle -1.6}}$ & 42.30\%$\textcolor{ForestGreen}{_{\scriptstyle +2.6}}$ & 7.30\%$\textcolor{BrickRed}{_{\scriptstyle -1.2}}$ & 21.90\%$\textcolor{ForestGreen}{_{\scriptstyle +1.4}}$ & 12.20$\textcolor{BrickRed}{_{\scriptstyle -1.0}}$ & 3.03\%$\textcolor{ForestGreen}{_{\scriptstyle +1.4}}$ \\
\rowcolor{lightgreen} %
\rowcolor{lightgreen} 
\qquad + native agentic traj. & 68.21\%$\textcolor{ForestGreen}{_{\scriptstyle +8.3}}$ & 55.81\%$\textcolor{ForestGreen}{_{\scriptstyle +12.1}}$ & 18.75\%$\textcolor{ForestGreen}{_{\scriptstyle +3.3}}$ & 47.59\%$\textcolor{ForestGreen}{_{\scriptstyle +7.9}}$ & 4.23\%$\textcolor{BrickRed}{_{\scriptstyle -4.3}}$ & 12.95\%$\textcolor{BrickRed}{_{\scriptstyle -7.6}}$ & 6.80$\textcolor{BrickRed}{_{\scriptstyle -6.4}}$ & 1.12\%$\textcolor{BrickRed}{_{\scriptstyle -0.6}}$ \\
\rowcolor{lightgreen} 
\qquad + context grounded traj.  & 62.73\%$\textcolor{ForestGreen}{_{\scriptstyle +2.8}}$ & 50.26\%$\textcolor{ForestGreen}{_{\scriptstyle +6.5}}$ & 21.00\%$\textcolor{ForestGreen}{_{\scriptstyle +5.5}}$ & 44.66\%$\textcolor{ForestGreen}{_{\scriptstyle +4.9}}$ & 8.99\%$\textcolor{ForestGreen}{_{\scriptstyle +0.5}}$ & 21.94\%$\textcolor{ForestGreen}{_{\scriptstyle +1.4}}$ & 8.46$\textcolor{BrickRed}{_{\scriptstyle -4.7}}$ & 1.12\%$\textcolor{BrickRed}{_{\scriptstyle -0.6}}$ \\
\rowcolor{lightgreen}

\quad \bf \midtooldata & \textbf{66.38\%}$\textcolor{ForestGreen}{_{\scriptstyle +6.4}}$ & \textbf{57.74\%}$\textcolor{ForestGreen}{_{\scriptstyle +14.0}}$ & \textbf{26.63\%}$\textcolor{ForestGreen}{_{\scriptstyle +11.1}}$ & \textbf{50.25\%}$\textcolor{ForestGreen}{_{\scriptstyle +10.5}}$ & \textbf{12.23\%}$\textcolor{ForestGreen}{_{\scriptstyle +3.7}}$ & \textbf{28.06\%}$\textcolor{ForestGreen}{_{\scriptstyle +7.6}}$ & \textbf{18.66}$\textcolor{ForestGreen}{_{\scriptstyle +5.5}}$ & \textbf{5.03\%}$\textcolor{ForestGreen}{_{\scriptstyle +3.4}}$ \\
\bottomrule
\end{tabular}%
}
\end{table*}

\textbf{General tool-use mid-training reveals a meaningful capability boundary.}
As shown in Table \ref{tab:mcp-universe-results}, although \midtooldata~substantially improves the overall score, the web search subset stays at 0.00. This contrast is informative rather than merely negative: other MCP domains, especially browser automation, financial analysis, and location, do improve meaningfully, so the issue is not a failure to transfer to MCP tools. Instead, it isolates a distinct class of agentic behavior, namely deep-search-style tasks that require longer-horizon evidence gathering, iterative refinement, and stronger agent-level control flow. This is an important empirical signal about the structure of agentic capability: broad tool-use supervision teaches reusable priors for schema grounding, tool selection, and interaction with unfamiliar APIs, while highly exploratory domains appear to require dedicated trajectory data and training objectives. In this sense, \midtooldata~helps map the boundary between general tool use and specialized agency, providing guidance for future mid-training work on search-heavy and other specialized settings.
\subsection{Ablation Study}
To study the effect of data design, we fix the 4B SFT recipe and vary only the mid-training corpus.
We decompose \midtooldata~additively: the processed raw sources alone, each synthesis branch added on top of them, and the complete mixture. We compare against a matched-budget generic mid-training baseline, \texttt{Dolmino-20BT} \citep{olmo2025olmo3}, with no-mid-training setting as a reference.

\textbf{The two \midtool~subsets play different but complementary roles.} The processed raw sources alone already provide a positive standalone signal: with no synthesized data at all, they improve BFCLv3 overall by $+2.6$ and MCP-Universe Pass by $+1.4$ over no mid-training, and stay competitive with \texttt{Dolmino-20BT} on BFCL overall and $\tau^2$-Bench Pass@4 while transferring substantially better to MCP-Universe. We note that \texttt{Dolmino-20BT} is itself not a purely non-synthetic
baseline, as the released mixture also contains model-generated components.
The complete mixture substantially improves over no mid-training, and even the variant that adds only context-grounded augmentation remains well above both no mid-training and \texttt{Dolmino-20BT}.
Compared with \texttt{Dolmino-20BT}, this variant is only marginally weaker on the live subset, but stronger on non-live, multi-turn, and overall.
On $\tau^2$-Bench, it is similar: the context-grounded-only variant exceeds both no mid-training and \texttt{Dolmino-20BT}. Notably, the native agentic trajectory subset is the only subset that contains trajectories synthesized by proprietary models.
The fact that this variant still substantially outperforms no mid-training on BFCLv3 and $\tau^2$-Bench, and exceeds \texttt{Dolmino-20BT} on nearly all metrics, indicates that the gains of \midtooldata~are not solely driven by proprietary-model-synthesized trajectories. This branch is nonetheless still model-synthesized, with an open-weight teacher; the strictly non-distilled contribution is isolated by the filtered data split above.

The two branches also contribute asymmetrically.
Adding native agentic trajectories alone yields the larger BFCLv3 gain ($+7.9$ vs.\ $+4.9$ overall), indicating that executable trajectories are especially important for precise function calling.
Adding context-grounded augmentation alone is the stronger of the two on $\tau^2$-Bench and MCP-Universe, indicating that grounding-oriented supervision is especially important for transfer.
On BFCL non-live the native-only variant is even slightly ahead of the complete mixture ($68.21\%$ vs.\ $66.38\%$), but it gives up large margins on multi-turn, $\tau^2$-Bench, and MCP-Universe; combining both branches is the only configuration that improves over no mid-training on all eight metrics.
MCP-Universe makes the complementarity especially clear: both single-branch variants fall below no mid-training, yet still outperform \texttt{Dolmino-20BT}, indicating that each subset contributes a meaningful but incomplete agentic prior. More broadly, the relatively competitive BFCL result of \texttt{Dolmino-20BT}, contrasted with its much weaker transfer on $\tau^2$-Bench and MCP-Universe, suggests that generic instruction-following-style mid-training may help simple function-calling behavior, but transfers poorly to more agentic settings.

\section{Related Work}
\paragraph{Mid-training for Agentic Capability}
The advancing agentic capability of frontier models emphasizes that strong agentic behavior is not obtained from post-training solely, but from combining a strong pretrained base with additional large-scale capability shaping for long-horizon mid-training \citep{moonshotai2026kimik25, xiao2026mimo}. For example, the GLM team describes its efforts in mid-training on reasoning and agentic data for long-context settings and on repo-level code data for software engineering tasks \citep{zeng2025glm45,zai2026glm5}.
Tongyi DeepResearch proposes agentic mid-training with a scaled agentic data pipeline for deep-research foundations \citep{su2025scalingagentscontinualpretraining, team2025tongyi}. \citet{kimidev} and \cite{davincidev} propose SWE-centric pipelines for agentic mid-training data. While \citet{olmo2025olmo3} discloses a mid-training recipe, it primarily focuses on high-quality data in general domains, with a mixture that emphasizes reasoning tasks such as math and science. Different from prior work \citep{wang2025octothinker}, we construct an open-source mid-training corpus for agentic tool use through a dedicated pipeline.

\paragraph{Post-training Dataset for Tool Use}
Early work on tool-use data focuses on construction from tool definitions or model self-annotation. Toolformer \citep{schick2023toolformer} shows that language models can insert API calls into text with self-supervision, while ToolLLM \citep{qin2023toolllm} and APIGen \citep{liu2024apigen} expand this line to large collections of real APIs and verifiable function-calling instances. 
Later work moves from single-turn invocation to richer agent interaction trajectories. AgentBank \citep{song2024agentbank} collects diverse interaction traces for general agent fine-tuning, and ToolACE \citep{liu2025toolacewinningpointsllm} shows that large-scale, carefully curated function-calling supervision remains highly effective for post-training tool-use alignment.
More recent work emphasizes multi-turn, environment-grounded data synthesis at larger scale. APIGen-MT \citep{prabhakar2025apigen} generates multi-turn trajectories through simulated agent-human interaction, TOUCAN \citep{xu2025toucansynthesizing15mtoolagentic} synthesizes 1.5M tool-agentic examples from real-world MCP environments, and Simia \citep{li2025simulating} focuses on synthetic environments and further explores simulator-backed training data for interactive agents. 
These efforts substantially improve the availability of post-training supervision for tool calling and agent interaction. Our focus is complementary and parallel to this line of work: we study an open-source corpus and data pipeline for agentic \emph{mid-training}, which can provide a broader prior for tool-use behavior and in turn work synergistically with downstream post-training datasets and objectives.

\section{Conclusion}
In this work, we study whether general tool use benefits from dedicated mid-training rather than being left entirely to post-training.
We introduce \midtool, a scalable corpus-construction pipeline, and \midtooldata, a 20.3B-token mixture that pairs filtered source corpora with context-grounded augmentation and native agentic trajectories.
Across 4B and 8B base models, mid-training on \midtooldata~consistently improves downstream performance on BFCL, $\tau^2$-Bench, and MCP-Universe under both SFT and RL, with especially strong gains on multi-turn and interactive settings.
These gains are complementary to, rather than a substitute for, stronger post-training: the downstream recipe is held fixed throughout our experiments, and we view scaling post-training supervision and strengthening tool-use mid-training as two axes that should advance together.
 
Ablations show that these improvements stem from the structure of the mixture: removing either synthesis branch degrades performance, indicating that grounding-oriented and execution-oriented supervision are complementary.
Our results also surface a meaningful capability boundary: while \midtooldata~improves broad transfer to unseen tools and MCP domains, it contributes marginally to deep-search-style exploratory behaviors. This suggests that general tool-use mid-training provides reusable priors for schema grounding and workflow composition, whereas highly exploratory domains require dedicated trajectory data.
Future work includes scaling native trajectory collection, broadening tool ecosystem coverage, and constructing specialized mid-training mixtures for domain-specific agentic behaviors.

\bibliographystyle{abbrvnat}
\nobibliography*
\bibliography{ref}

\clearpage
\appendix
\section{More Details for \midtool Pipeline }
\subsection{Benchmark Exclusion During Data Collection}\label{appx:decontamination}
To reduce benchmark leakage from the GitHub repository slice, we explicitly exclude known benchmark and evaluation repositories before repository crawling and preprocessing. In practice, our code-data pipeline loads a blacklist of repository names and skips matched repositories entirely during dataset construction. The blacklist includes repositories associated with BFCL-related benchmarks, $\tau$-Bench / $\tau^2$-Bench, MCP-Universe and related MCP evaluation suites, together with other widely used agentic and tool-use benchmarks.

This blacklist-based exclusion is integrated into the data pipeline rather than applied as an ad hoc postprocessing step. We actively maintain and update this list as new public agentic benchmarks appear, so that future runs of the pipeline can continue to enforce the same benchmark-exclusion policy during data collection.

\subsection{Web and PDF Filtering Pipeline}\label{appx:web-pdf-pipeline}
For web and PDF midtraining sources, we use a separate four-phase pipeline consisting of keyword/URL filtering, fastText classification, quality filtering, and MinHash-based deduplication.

\paragraph{Phase 1: High-recall prescreening.}
The first phase acts as a high-recall prescreener. It combines matches against a curated vocabulary of software-development terms with URL patterns associated with documentation and technical reference sites, and it gives additional weight to documents containing code-like structure. This phase is intended to retain broadly relevant technical content, such as API references, SDK or library documentation, CLI-style help pages, platform documentation, troubleshooting pages, and developer Q\&A, while removing most general web noise.

\paragraph{Phase 2: fastText classification.}
The second phase is a lightweight fastText \citep{fasttext} classifier trained from LLM-labeled seed data. To build the seed set, we sample from 1M web documents and 3M PDF documents. We then use \texttt{Qwen2.5-7B-Instruct} to annotate the seed set. Our labeling prompt defines positive examples, including documentation, tutorials, configuration, debugging traces, and code-centered technical discussion, and negatives as non-technical or low-value material for tool-use purposes. These labeled examples are then converted into positive and negative training sets for fastText. We apply stricter classification thresholds for PDF data because PDF extraction is noisier and more heterogeneous than web text.
 
\paragraph{Phase 3: Quality filtering.}
The third phase applies document-level quality controls. For both web and PDF data, we filter on language confidence, length, word count, symbol density, and estimated code ratio. The goal is to remove low-quality extraction artifacts, extremely short or noisy documents, and documents dominated by uninterpretable symbols or raw code dumps. For PDF data, we additionally apply an OCR-quality filter to address extraction noise.

\paragraph{Phase 4: Deduplication.}
Finally, we use MinHash LSH to reduce repeated content across sources and shards.

\subsection{Post-hoc Contamination Analysis}\label{appx:decon}
The blacklist described above prevents benchmark repositories from entering the code slice, but it does not bound leakage that may arrive through web pages, PDFs, or model-synthesized supervision. We therefore additionally audit the finished mixture with DeCon~\citep{olmo2025olmo3}, using its default configuration and lowering the token-length threshold so that short benchmark items are also covered. We scan the web and PDF slices together with the teacher-synthesized data against all three evaluation benchmarks used in this paper: BFCLv3, $\tau^2$-Bench, and MCP-Universe.

DeCon flags fewer than 20 candidates in total, all of them from the web slice and all against BFCLv3, and none for $\tau^2$-Bench or MCP-Universe. Manual inspection confirms that every flagged item is a false positive: generic function-calling and API-documentation text that shares surface $n$-grams with benchmark items, but that contains no benchmark instances and no reference answers. We therefore find no evidence of actual benchmark leakage into \midtooldata.

Since DeCon operates by surface-level $n$-gram matching, this analysis bounds verbatim overlap rather than semantic or schema-level similarity, which is particularly relevant for MCP-related tool definitions. We leave a deeper semantic audit of such overlap to future work.

\subsection{Composition Analysis of \midtooldata}\label{appx:composition}
Table~\ref{tab:data-mixture} in the main text reports the mixture at the level of source families. Here we provide a finer-grained view of what each slice actually contains.

\paragraph{Augmentation type per source.}
A source-level breakdown alone hides how much of each slice carries synthesized supervision. Decomposing every \texttt{web}/\texttt{pdf}/\texttt{code} sample into one of \{source-only, $+$QA, $+$QA $+$ trajectory\} gives Table~\ref{tab:appx-aug-type}. The three slices play visibly different roles: web is dominated by QA-style augmentation, PDF contributes the largest share of full trajectories, and code remains mostly raw source, reflecting that much of its tool-use signal is already present in the original files.

\begin{table}[h]
\centering
\caption{Augmentation type per source slice, as a fraction of samples within that slice. Sample counts match the \emph{Samples} column of Table~\ref{tab:data-mixture}.}
\label{tab:appx-aug-type}
\small
\begin{tabular}{lccc}
\toprule
Slice & source-only & $+$QA & $+$QA $+$ traj. \\
\midrule
web (6.86M)  & 36.4\% & 52.2\% & 11.3\% \\
pdf (1.34M)  & 30.8\% & 13.4\% & 55.7\% \\
code (2.60M) & 68.9\% & 4.2\%  & 26.8\% \\
\bottomrule
\end{tabular}
\end{table}

\paragraph{Trajectory shape.}
Among samples that contain agentic trajectories, Table~\ref{tab:appx-traj-stats} reports per-document statistics, where a turn is a user message, a step is an assistant message, and a tool call is a single tool invocation. Context-grounded trajectories derived from documents tend to have more assistant steps and tool calls per user turn, whereas native trajectories synthesized from executable environments are longer in user turns.

\begin{table}[h]
\centering
\caption{Per-document trajectory statistics among samples containing agentic trajectories. A turn is a user message, a step is an assistant message, and a tool call is a single tool invocation.}
\label{tab:appx-traj-stats}
\small
\begin{tabular}{llcccc}
\toprule
Metric & Stat & web & pdf & code & native \\
\midrule
\multirow{3}{*}{Turns} & mean & 2.8 & 3.0 & 3.1 & 4.0 \\
 & median & 3 & 3 & 3 & 4 \\
 & max & 15 & 17 & 8 & 69 \\
\midrule
\multirow{3}{*}{Steps} & mean & 5.3 & 7.1 & 6.0 & 3.8 \\
 & median & 5 & 6 & 5 & 4 \\
 & max & 54 & 77 & 58 & 21 \\
\midrule
\multirow{3}{*}{Tool calls} & mean & 3.1 & 4.8 & 3.3 & 2.4 \\
 & median & 2 & 4 & 2 & 2 \\
 & max & 72 & 78 & 103 & 66 \\
\bottomrule
\end{tabular}
\end{table}

\paragraph{Tool inventory.}
\midtooldata~exposes 2.60M unique tool names across its tool calls. Mapping names to functional categories with keyword rules gives Table~\ref{tab:appx-tool-inventory}. Beyond the head categories, 37.2\% of all calls fall into a domain-specific long tail, which quantifies the tool diversity of the mixture and distinguishes it from prior tool-use post-training corpora that are typically built over a fixed tool inventory.

\begin{table}[h]
\centering
\caption{Functional categories of tool calls in \midtooldata, obtained by keyword mapping over tool names.}
\label{tab:appx-tool-inventory}
\small
\begin{tabular}{lrrl}
\toprule
Category & Calls & Share & Examples \\
\midrule
Code execution / Shell / DevOps & 862K & 10.9\% & \texttt{run\_command}, \texttt{kubectl\_get} \\
List / Get (generic)       & 811K  & 10.2\% & \texttt{list\_tools}, \texttt{get\_*} \\
File / Filesystem          & 380K  & 4.8\%  & \texttt{read\_file}, \texttt{list\_dir} \\
Set / Update / Modify      & 354K  & 4.5\%  & \texttt{set\_*}, \texttt{configure} \\
Add / Create               & 348K  & 4.4\%  & \texttt{create\_*}, \texttt{register\_*} \\
Search / Lookup            & 335K  & 4.2\%  & \texttt{search}, \texttt{lookup} \\
AI / NLP / ML              & 249K  & 3.1\%  & \texttt{BertFor*}, \texttt{embed} \\
Web / HTTP / API           & 243K  & 3.1\%  & \texttt{http\_get}, \texttt{fetch\_url} \\
Database / SQL             & 201K  & 2.5\%  & \texttt{sql\_query}, \texttt{select\_*} \\
Hardware / Sensor / Device & 153K  & 1.9\%  & \texttt{read\_register}, \texttt{modbus\_*} \\
13 other categories        & 920K  & 11.6\% & Maps, Auth, Email, Finance, E-commerce \\
Domain-specific long tail  & 2.96M & 37.2\% & \texttt{formNavigator}, \texttt{MemcachedSet} \\
\bottomrule
\end{tabular}
\end{table}

\section{Experimental Details}\label{appx:hyperparameters}

We summarize the main training hyperparameters used in our experiments here.
As described in the main text, both mid-training and SFT are conducted with ArcticTraining on 32 H200 GPUs, while RL follows the AWM setup with VeRL-style GRPO training on 8 B200 GPUs.
Dataset composition details is described in the main text.
The critical hyperparameters for mid-training, SFT, and RL are listed in Table~\ref{tab:appx-mt-hparams}, Table~\ref{tab:appx-sft-hparams}, and Table~\ref{tab:appx-rl-hparams}, respectively.

\begin{table}[h]
\centering
\caption{Mid-training hyperparameters for \texttt{Qwen3-4B-Base} and \texttt{Qwen3-8B-Base}.}
\label{tab:appx-mt-hparams}
\small
\begin{tabular}{ll}
\toprule
Hyperparameter & Value \\
\midrule
Epochs & 1 \\
Max Sequence length & 8192 \\
Optimizer & AdamW \\
Learning rate & $3\times 10^{-5}$ \\
Betas & $(0.9, 0.999)$ \\
Weight decay & 0.01 \\
LR schedule & WSD \\
Warmup steps & 50 \\
Target global token batch & 4M tokens (with data packing)  \\
\bottomrule
\end{tabular}
\end{table}

\begin{table}[h]
\centering
\caption{Supervised fine-tuning hyperparameters.}
\label{tab:appx-sft-hparams}
\small
\begin{tabular}{ll}
\toprule
Hyperparameter & Value \\
\midrule
Max Sequence length & 32768 \\
Optimizer & AdamW \\
Learning rate & $2\times 10^{-5}$ \\
Betas & $(0.9, 0.95)$ \\
Weight decay & 0.01 \\
LR schedule & cosine, warmup ratio 0.001 \\
Training batch size & 128 \\
\bottomrule
\end{tabular}
\end{table}

\begin{table}[h]
\centering
\caption{Reinforcement-learning hyperparameters.}
\label{tab:appx-rl-hparams}
\small
\begin{tabular}{ll}
\toprule
Hyperparameter & Value \\
\midrule
RL algorithm & GRPO \\
Total training steps & 64 \\
Learning rate & 4B: $1\times 10^{-6}$; 8B: $5\times 10^{-7}$ \\
Data Batch size & 32 \\
PPO Mini-batch size & 32 \\
Rollouts per prompt & 16 \\
KL coefficient & 0.001 \\
Entropy coefficient & 0.0 \\
Clip ratio (high) & 0.28 \\
History limit & 3 \\
Maximum agent turns & 20 \\
\bottomrule
\end{tabular}
\end{table}

\section{More Experimental Analysis}\label{appx:exp-result}
\subsection{SFT Convergence Analysis}\label{appx:sft-convergence}
We further analyze the SFT training process using the same comparative-loss perspective as prior work on agentic continual pre-training \citep{su2025scalingagentscontinualpretraining}.
Here, we ask whether \midtooldata~produces a better initialization for downstream agentic post-training than either the raw base model or a generic continued-pretraining baseline.

\begin{figure*}[t]
\centering
\includegraphics[width=0.8\textwidth]{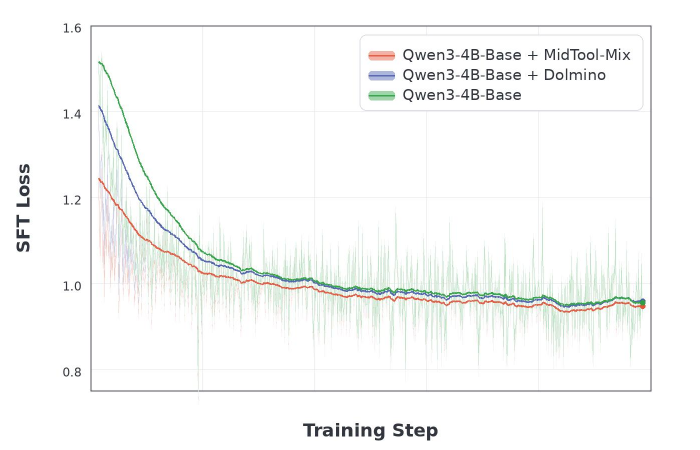}
\caption{SFT loss trajectories on the same downstream tool-use corpus for three Qwen3-4B initializations.
\midtooldata~starts from a lower loss, converges faster, and maintains the best loss throughout training compared with both the raw base model and the Dolmino mid-trained baseline.}
\label{fig:appx-sft-loss}
\end{figure*}

We compare three 4B initializations under the same SFT setup and downstream corpus: \texttt{Qwen3-4B-Base}, \texttt{Qwen3-4B-Base} continued on Dolmino, and \texttt{Qwen3-4B-Base} continued on \midtooldata.
Figure~\ref{fig:appx-sft-loss} reports the training loss trajectories over SFT steps, including both the raw per-step loss and a smoothed trend curve for readability.
Because the downstream data, optimizer settings, and training recipe are held fixed, differences in convergence primarily reflect the quality of the initialization induced by the preceding mid-training stage.

\paragraph{Mid-training improves post-training optimization efficiency.}
Figure~\ref{fig:appx-sft-loss} shows the SFT training loss.
Among the three initializations, \midtooldata~enters SFT with the lowest loss, descends fastest in the early stage, and maintains the best loss throughout nearly the entire training run.
This pattern matters because all three models are fine-tuned on the same downstream corpus with the same optimizer and schedule, so the gap is best explained by the quality of the initialization rather than by differences in the post-training recipe itself.
We note, however, that SFT loss measures next-token prediction on the downstream corpus and is not itself a measure of tool-use capability: initializations that reach a similar loss can still differ substantially in schema grounding, tool selection, and multi-turn execution~\citep{liu2023same, isik2025scaling}. We therefore read these curves as evidence about optimization behavior, and rely on the main-text benchmark results for capability claims.

The comparison against Dolmino is especially informative.
Generic midtraining does improve over the raw base model, but it still converges more slowly and to a worse loss than \midtooldata.
This suggests that the benefit is not merely a consequence of extra training tokens or additional compute.
Instead, the agentic and tool-centered composition of \midtooldata~appears to provide a better inductive bias for downstream tool-use supervision, making subsequent SFT and RL easier to optimize.
This observation is consistent with the main-text results: the models that perform best after post-training are also the ones that begin SFT from a more favorable optimization landscape.

\subsection{RL Reward Analysis}\label{appx:rl-reward}
We next examine the RL training dynamics to understand whether the gains from \midtooldata~also persist during reinforcement learning.
Here, the central question is not only the final reward reached in the training environment, but also how efficiently the model adapts during RL and whether that in-environment improvement translates to broader downstream generalization.

\begin{figure*}[t]
\centering
\begin{minipage}[t]{0.49\textwidth}
\centering
\includegraphics[width=\linewidth]{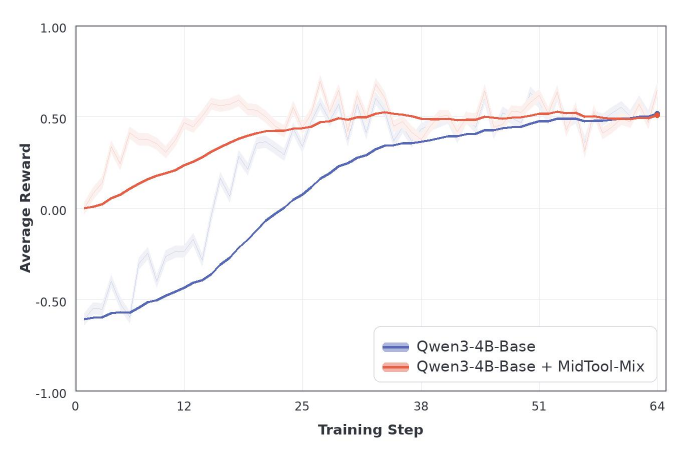}
\end{minipage}
\hfill
\begin{minipage}[t]{0.49\textwidth}
\centering
\includegraphics[width=\linewidth]{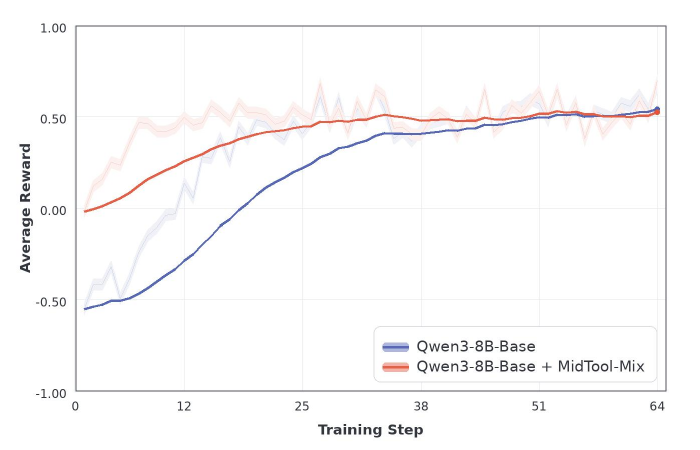}
\end{minipage}
\caption{Average RL reward during training for 4B (left) and 8B (right) models.
In both scales, \midtooldata~starts from a stronger point and improves more quickly in the early stage of RL, while the non-mid-trained baseline gradually catches up later within the same training environment.}
\label{fig:appx-rl-reward}
\end{figure*}

\paragraph{Mid-trained models adapt faster in RL, while their main advantage appears in out-of-environment generalization.}
Figure~\ref{fig:appx-rl-reward} shows a consistent pattern across both 4B and 8B.
The \midtooldata~models begin RL from clearly higher reward and improve much more rapidly in the early stage, whereas the corresponding base models start from substantially lower reward and require many more steps to reach a similar level.
This suggests that mid-training provides a more favorable initialization not only for SFT, but also for subsequent policy optimization, allowing the model to adapt more efficiently once RL begins.

At the same time, the reward curves also show an important limitation of in-environment training reward as a proxy for final agent quality.
By later RL steps, the two curves move much closer at both model sizes, which indicates that the raw base model can eventually learn much of the policy required for this specific RL environment.
But this narrowing gap does not imply that the two training pipelines are equivalent.
In the main results, the mid-trained models still achieve clearly stronger performance across BFCL, $\tau^2$-Bench, and MCP-Universe after RL, even when the in-training rewards become similar.
Taken together, these results suggest that RL reward primarily reflects adaptation to the training environment, so its convergence does not imply that the two pipelines behave equivalently downstream. Our evidence for this is the persisting post-RL benchmark gap rather than the reward curves themselves.
In other words, the main value of mid-training is not just faster reward acquisition during RL, but stronger generalizability after RL.

\subsection{Pilot Study on Visual Tool Use}\label{appx:visual-tool-use}
Inspired by the discussion of Kimi K2.5 \citep{moonshotai2026kimik25}, we also run a small pilot study on visual tool use as a zero visual tool cold-start transfer setting.
This experiment is not the focus of our paper, rather we view it as an exploratory but potentially informative result: we simply extend text-only tool-use training and ask whether any capability transfers to a visual tool-use.
Similar to our main setting, we mid-train a multimodal base model, \texttt{gemma-3-4b-pt}, on \midtooldata. For both the mid-trained model and the original base model, we then fine-tune it on 37.5k samples from FineVision \citep{wiedmann2025finevision}, followed by the same text-based tool SFT as in our main setting. For both mid-training and SFT, we do not include any visual tool-use data in training.

We evaluate on VisualToolBench \citep{visualtoolbench} using single-turn subset due to its complexity. As the benchmark only partially releases the evaluation harness, we adapt our own implementation of the evaluation stack.
We report the tool success rate and the benchmark's average rubric score (ARS).

\begin{table*}[!b]
\centering
\caption{Exploratory results on the single-turn subset of VisualToolBench.
We fine-tune the base model on a FineVision subset and then on the same text-only tool-use SFT dataset as in our main setting.
We report tool success rate and average rubric score; higher is better. \texttt{Llama4-Maverick} score is reported from the original benchmark paper for reference.}
\label{tab:appx-visualtoolbench-results}
\small
\resizebox{\textwidth}{!}{%
\begin{tabular}{lccccccc}
\toprule
Model & Tool Succ. & Overall & STEM & Med & Fin & Sprt & Gen \\
\midrule
\texttt{Llama4-Maverick}$^*$ & - & 0.1545 &  0.1875 &  0.1581 &  0.1182 &  0.1562 &  0.1524 \\
\texttt{Gemma3-4B-pt} + SFT & 0.5863 & 0.0567 & 0.0655 & 0.0600 & 0.0573 & 0.0564 & 0.0448 \\
\texttt{Gemma3-4B-pt} + \midtooldata + SFT & 0.7231 &  0.0661 & 0.0733 & 0.0765 & 0.0640 & 0.0563 & 0.0606 \\
\bottomrule
\end{tabular}%
}
\end{table*}

\paragraph{A small but striking transfer signal.}
Despite the difficulty of the benchmark, the mid-trained checkpoint shows a consistent agentic advantage over the baseline.
It invokes tools more often, executes tools with higher success rate, and achieves marginally better single-turn performance across most grouped domains.
Concretely, adding \midtooldata~improves the tool success rate from 0.5863 to 0.7231, and it also improves the overall rubric score from 0.0567 to 0.0661, with gains in most domains.
We therefore view this result as an explosive early signal that general tool-use mid-training may transfer beyond the text-only setting, even without vision-centric tool use pretraining and post-training.

\paragraph{Lessons for building stronger agentic intelligence.}
First, in a substantial portion of VisualToolBench, successful tool use requires environment-oriented code fullfill for visual tool invocation (e.g., image processing), and failures often arise at this stage.
This code-heavy form of tool use is meaningfully different from the more schema-grounded function-calling and API-style tool use emphasized by our main setting, but is central to CLI-style agency and SWE-oriented agency.
Second, even when the model successfully calls a tool, it does not always incorporate the returned evidence into the final answer.
This pattern resembles what we observe for web-search tasks in MCP-Universe: the model can trigger the tool, but does not always fully ground the final response in the tool outputs.
Taken together, these observations suggest a natural future direction: augmenting mid-training with more code-heavy agentic tool-use data, especially for execution-centric environments, and developing training signals that more explicitly teach the model to ground final answers in tool results, including redundant or partially overlapping evidence returned by multiple tool interactions.

\subsection{Illustrative Examples}\label{appx:data-example}
\begin{figure*}[t]
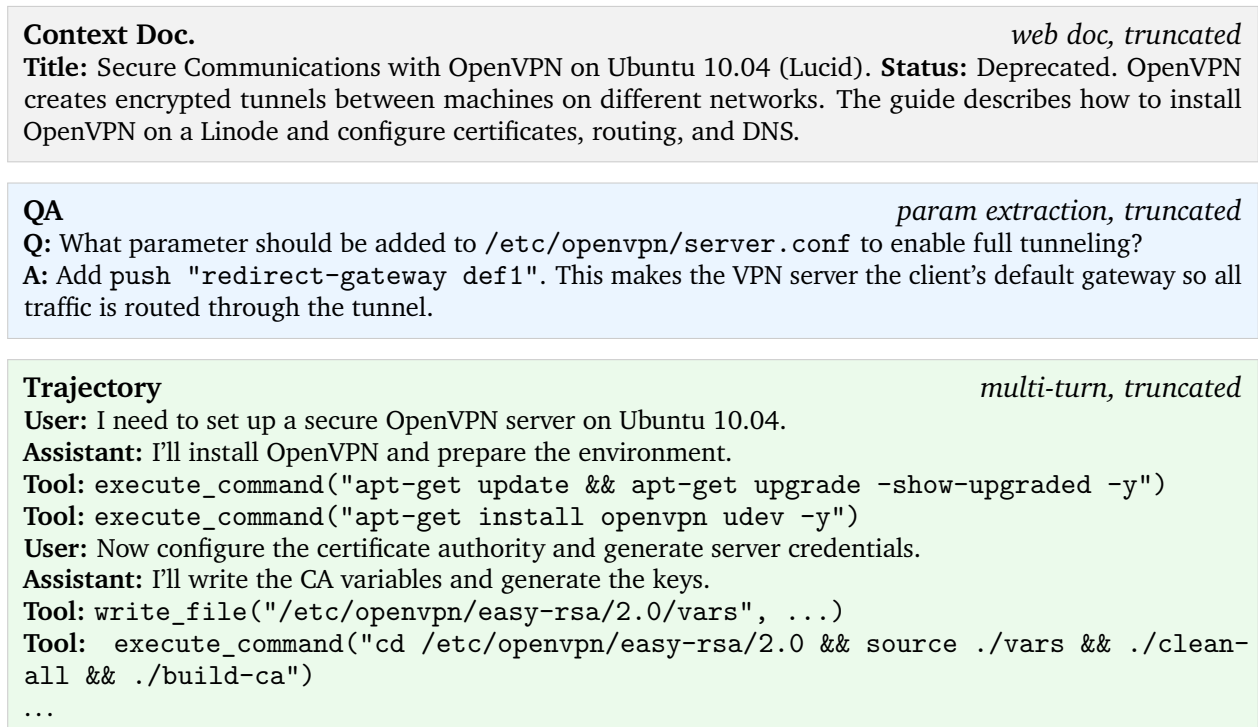

\centering
\setlength{\fboxsep}{6pt}

\fcolorbox{black!20}{lightgray}{%
\parbox{0.97\textwidth}{%
\textbf{Context Doc.} \hfill \textit{web doc, truncated}

\small
\textbf{Title:} Secure Communications with OpenVPN on Ubuntu 10.04 (Lucid).
\textbf{Status:} Deprecated.
OpenVPN creates encrypted tunnels between machines on different networks.
The guide describes how to install OpenVPN on a Linode and configure certificates, routing, and DNS.
}}%

\vspace{0.6em}

\fcolorbox{black!20}{midtool}{%
\parbox{0.97\textwidth}{%
\textbf{QA} \hfill \textit{param extraction, truncated}

\small
\textbf{Q:} What parameter should be added to \texttt{/etc/openvpn/server.conf} to enable full tunneling?

\textbf{A:} Add \texttt{push "redirect-gateway def1"}.
This makes the VPN server the client's default gateway so all traffic is routed through the tunnel.
}}%

\vspace{0.6em}

\fcolorbox{black!20}{lightgreen}{%
\parbox{0.97\textwidth}{%
\textbf{Trajectory} \hfill \textit{multi-turn, truncated}

\small
\textbf{User:} I need to set up a secure OpenVPN server on Ubuntu 10.04.

\textbf{Assistant:} I'll install OpenVPN and prepare the environment.

\textbf{Tool:} \texttt{execute\_command("apt-get update \&\& apt-get upgrade --show-upgraded -y")}

\textbf{Tool:} \texttt{execute\_command("apt-get install openvpn udev -y")}

\textbf{User:} Now configure the certificate authority and generate server credentials.

\textbf{Assistant:} I'll write the CA variables and generate the keys.

\textbf{Tool:} \texttt{write\_file("/etc/openvpn/easy-rsa/2.0/vars", ...)}

\textbf{Tool:} \texttt{execute\_command("cd /etc/openvpn/easy-rsa/2.0 \&\& source ./vars \&\& ./clean-all \&\& ./build-ca")} \\
\ldots
}}%

\caption{Illustrative training datum from the web slice.
A single example contains (top) a context documentation, (middle) a QA pair derived from that source, and (bottom) a multi-turn trajectory.
For readability, long spans are truncated and tool calls are lightly normalized.}
\label{fig:data-example-openvpn}
\end{figure*}

We show two representative Examples (truncated for space).
Figure~\ref{fig:data-example-openvpn} is a web-derived triplet containing source text, a QA pair, and a tool-augmented trajectory.
Figure~\ref{fig:data-example-tappi} is a native agentic trajectory stored as a single flattened conversation with interleaved tool calls and responses.


\begin{figure*}[t]
\centering
\setlength{\fboxsep}{6pt}

\fcolorbox{black!20}{lightgreen}{%
\parbox{0.97\textwidth}{%
\raggedright
\textbf{Native agentic trajectory} \hfill \textit{ truncated}

\small
\textbf{System:} You are a helpful assistant with access to the following tools.
\textbf{Tools:} \texttt{tappi.open}, \texttt{tappi.elements}, \texttt{tappi.click}, \texttt{tappi.type},
\texttt{tappi.text}, \texttt{tappi.screenshot}, \ldots

\textbf{User:} Reproduce a checkout bug on \url{https://shop.mountaintrailsupply.com/} where the ``Place order'' button spins forever.

\textbf{Assistant:} I will open the storefront, follow the customer's steps, and capture any failure text or screenshot.
\textbf{Tool:} \texttt{tappi.open(storefront)} \\

\textbf{User:} A direct product URL is available; use it, add one item to cart, and confirm the cart text.
\textbf{Tool:} \texttt{tappi.open(product\_url)} \\
\textbf{Tool:} \texttt{tappi.elements()} \\
\textbf{Tool:} \texttt{tappi.click(index=2)} \\
\textbf{Tool:} \texttt{tappi.text()}  \\

\textbf{Assistant:} The item is in cart; next I will proceed to checkout, fill dummy data, click ``Place order'', then inspect the stuck state.
\textbf{Tool:} \texttt{tappi.click(index=3)} \\
\textbf{Tool:} \texttt{tappi.type(index=0, text=email)} \\
\textbf{Tool:} \texttt{tappi.type(index=5, text=card)} \\
\textbf{Tool:} \texttt{tappi.click(index=8)} \\
\textbf{Tool:} \texttt{tappi.text()} \\
\textbf{Tool:} \texttt{tappi.screenshot(path=artifacts/checkout-spinner.png)}

\textbf{Tool response:} ``Checkout \ldots Processing your order\ldots Please wait while we complete your order \ldots Place order'' \\
\textbf{Tool response:} screenshot saved to \texttt{artifacts/checkout-spinner.png}

\textbf{Assistant:} The page remained on \url{https://shop.mountaintrailsupply.com/checkout} with the visible message
``Processing your order\ldots'' and
``Please wait while we complete your order \ldots''.
This matches the customer's report that the checkout button spins indefinitely.
}}%

\caption{Illustrative native agentic trajectory example. Long spans are truncated and lightly normalized for readability.}
\label{fig:data-example-tappi}
\end{figure*}

\section{Limitations and Future Work}\label{appx:limitations}
\midtool~is a first step toward treating general tool use as a mid-training problem, and the questions it leaves open are, in our view, the most interesting part of this direction.

\paragraph{Co-designing mid-training and post-training.}
Our experiments deliberately hold the downstream recipe fixed so that the effect of the mid-training corpus can be isolated, and our compute budget did not allow us to also sweep study the post-training axis. The relationship between the two stages is therefore largely unexplored, and we believe it is the most promising next step: how the composition of the SFT corpus interacts with the mid-training mixture, whether stronger or larger post-training supervision substitutes for or compounds with a tool-use prior, and how the choice of RL environments and reward design shifts what the prior is worth. Each of these deserves its own ablation, and we regard them as core questions for mid-training research rather than as details of the present study.

\paragraph{Mapping the mixture design space at matched budget.}
The ablation in Table~\ref{tab:ablation-bfcl} varies the mixture under a fixed recipe, which already separates raw source signal from synthesized supervision. A fuller account would hold the token budget constant across variants and add controls such as synthesized QA pairs without their originating context, an equal-compute mixture of generic technical documentation, and a native-trajectory-only mixture scaled to the full budget. The last of these is a substantial undertaking on its own, since it requires executable tool environments, interaction generation, and validation at the scale of tens of billions of tokens, and we see it as an interesting systems-and-data challenge in its own right.

\paragraph{Reducing dependence on strong teachers.}
Both synthesis branches of \midtool~currently rely on teacher models. As open-weight agentic models continue to improve, the same pipeline can be instantiated with progressively smaller or self-generated teachers, and the point at which a model can usefully synthesize its own tool-use mid-training data becomes an informative measure of agentic maturity. We expect this question to be most meaningful at larger student scales, where self-generation is diverse and reliable enough to sustain the pipeline's validation requirements.

\paragraph{Beyond general tool use.}
Finally, our results locate a boundary rather than a ceiling: a general tool-use prior transfers broadly across unfamiliar schemas and MCP domains, while search-heavy and other exploratory behaviors appear to require their own supervision. Extending the mid-training view to deep search, software engineering, and vertical agentic workflows, and understanding which priors are shared across them, is the broader agenda we hope this work helps open.

\end{document}